\documentclass[12pt]{article}
\usepackage[T1]{fontenc}
\usepackage{graphicx}

\begin{document}

\bibliographystyle{authordate2}

\title{ADAPTIVE LEARNING WITH BINARY NEURONS}

\author{Juan-Manuel Torres-Moreno$^1$ and Mirta B. Gordon$^2$\\
$^1$ Laboratoire Informatique d'Avignon\\
BP 1228 84911 Avignon Cedex 09, France\\
$^2$ Laboratoire TIMC-IMAG (UMR CNRS-UJF 5525)\\
Domaine de La Merci - Bât. Jean Roget 38706 La Tronche, France\\
}


\maketitle

\begin{abstract}
\noindent A efficient incremental learning algorithm for classification tasks,
called NetLines, well adapted for both binary and real-valued input
patterns is presented. It generates small compact feedforward
neural networks with one hidden layer of binary units and binary
output units. A convergence theorem ensures that solutions with a
finite number of hidden units exist for both binary and real-valued
input patterns. An implementation for problems with more than two
classes, valid for any binary classifier, is proposed. The
generalization error and the size of the resulting networks are
compared to the best published results on well-known classification
benchmarks. Early stopping is shown to decrease overfitting,
without improving the generalization performance.
\end{abstract}


\section{Introduction}

Feedforward neural networks have been successfully applied to the
problem  of learning from examples pattern classification. The
relationship between number of weights, learning capacity and
network's generalization ability is well understood only for the
simple perceptron, a single binary unit whose output is a sigmoidal
function of the weighted sum of its inputs. In this case, efficient
learning algorithms based on theoretical results allow the
determination of the optimal weights. However, simple perceptrons
can only generalize those (very few) problems in which the input
patterns are {\it linearly separable} (LS). In many actual
classification tasks, multilayered perceptrons with hidden units
are needed. However, neither the architecture (number of units,
number of layers) nor the functions that hidden units have to learn
are known {\it a priori}, and the theoretical understanding of
these networks is not enough to provide useful hints.

Although pattern classification is an intrinsically discrete task,
it may be casted as a problem of {\it function approximation} or
{\it regression}, by assigning real values to the targets. This is
the approach used by Backpropagation and related algorithms, which
minimize the squared training error of the output units. The
approximating function must be highly non-linear, as it has to fit
a constant value inside the domains of each class, and present a
large variation at the boundaries between classes. For example, in
a binary classification task in which the two classes are coded as
$+1$ and $-1$, the approximating function must be constant and
positive in the input space regions or domains corresponding to
class $1$, and constant and negative for those of class $-1$. The
network's weights are trained to fit this function everywhere, in
particular inside the class-domains, instead of concentrating on
the relevant problem of the determination of the frontiers between
classes. As the number of parameters  needed for the fit is not
known {\it a priori}, it is tempting to train a large number of
weights, that allow to  span, at least in principle, a large set of
functions which is expected to contain "true" one. This introduces
a small bias\cite{nc:Geman+Bienenstock+Doursat:1992}, but leaves us
with the difficult problem of minimizing a cost function in a high
dimensional space, with the risk that the algorithm gets stuck in
spurious local minima, whose number grows with the number of
weights. In practice, the best generalizer is determined through a
trial and error process in which both the number of neurons and
weights are varied.

An alternative approach is provided by {\it incremental}, {\it
adaptive} or {\it growth} algorithms, in which the hidden units are
successively added to the network. One advantage is fast learning,
not only because the problem is reduced to training simple
perceptrons, but also because adaptive procedures do not need the
trial and error search for the most convenient architecture. Growth
algorithms allow  the use of {\it binary} hidden neurons, well
suited for building hardware dedicated devices. Each binary unit
determines a domain boundary in input space. Patterns lying on
either side of the boundary are given different hidden states.
Thus, all the patterns inside a domain in input space are mapped to
the same {\it internal representation} (IR). This binary encoding
is different for each domain. The output unit performs a logic
(binary) function of these IRs, a feature that may be useful for
rule extraction. As there is not a unique way of associating IRs to
the input patterns, different incremental learning algorithms
propose different targets to be learnt by the appended hidden
neurons. This is not the only difference: several heuristics exist
that generate fully connected feedforward networks with one or more
layers, and tree-like architectures with different types of neurons
(linear, radial basis functions). Most of these algorithms are not
optimal with respect to the number of weights or hidden units.
Indeed, growth algorithms have often been criticized because they
may generate too large networks, generally believed to be bad
generalizers because of overfitting.

The aim of this paper is to present a new incremental learning
algorithm for binary classification tasks, that generates small
feedforward networks. These networks have a single hidden layer of
{\it binary} neurons fully connected to the inputs, and a single
output neuron connected to the hidden units. We propose to call it
{\it NetLines}, for {\bf N}eural {\bf E}ncoder {\bf T}hrough {\bf
Line}ar {\bf S}eparations. During the learning process, the targets
that each appended hidden unit has to learn help to decrease the
number of classification errors of the output neuron. The crucial
test for any learning algorithm is the generalization ability of
the resulting network. It turns out that the networks built with
NetLines are generally smaller, and generalize better, than the
best networks found so far on well-known benchmarks. Thus, large
networks do not necessarily follow from growth heuristics. On the
other hand, although smaller networks may be generated with
NetLines through early stopping, we found that they do not
generalize better than the networks that were trained until the
number of training errors vanished. Thus, overfitting does not
necessarily spoil the network's performance. This surprising result
is in good agreement with recent work on the bias/variance dilemma
\cite{Bias_Variance_Curse} showing that, unlike in regression
problems where bias and variance compete in the determination of
the optimal generalizer, in the case of classification they combine
in a highly non linear way.

Although NetLines creates networks for two-class problems,
multi-class problems may be solved using any strategy that combines
binary classifiers, like winner-takes-all. In the present work we
propose a more involved approach, through the construction of a
{\it tree of networks}, that may be coupled with any binary
classifier.

NetLines is an efficient approach to create small compact
classifiers for problems with binary or continuous inputs. It is
most suited for problems where a discrete classification decision
is required.  Although it may estimate posterior probabilities, as
discussed in section \ref{sec:interpretation}, it requires more
information than the bare network's output. Another weakness of
NetLines is that it is not simple to retrain the network when, for
example, new patterns are available or class priors change over
time.

The paper is organized as follows: in section \ref{sec:Learning} we
give the basic definitions and present a simple example of our
strategy. This is followed by the formal presentation of the growth
heuristics and the perceptron learning algorithm used to train the
individual units. In section \ref{sec:Comparison} we compare
NetLines to other growth strategies. The construction of trees of
networks for multi-class problems is presented in section
\ref{sec:Multiclass}. A comparison of the generalization error and
the network's size, with respect to results obtained with other
learning procedures, is presented in section \ref{sec:Benchmarks}.
The conclusions are left to section \ref{sec:conclusion}.

\section{The Incremental Learning Strategy}
\label{sec:Learning}

\subsection {Definitions}

We first present our notation and basic definitions. We are given
a training set of $P$ input-output examples $\{\vec{\xi}^{\mu},
\tau^{\mu}$\}, where $\mu=1,2,\cdots,P$. The inputs $\vec{\xi}^{\mu}=(1,\xi_1^{\mu},
\xi_2^{\mu}, \cdots, \xi_N^{\mu})$
may be binary or real valued $N+1$ dimensional vectors. The first
component $\xi_0^{\mu} \equiv 1$, the same for all the patterns,
allows to treat the bias as a supplementary weight. The outputs are
binary, $\tau^{\mu}=\pm 1$. These patterns are used to learn the
classification task with the growth algorithm. Assume that, at a
given stage of the learning process, the network has already $h$
binary neurons in the hidden layer. These neurons are connected to
the $N+1$ input units through synaptic weights $\vec{w}_k=(w_{k0},
w_{k1}
\cdots w_{kN})$, $1 \leq k \leq h$, $w_{k0}$ being the bias.

Then, given an input pattern $\vec{\xi}$, the states $\sigma_k$ of
the  hidden neurons ($1 \leq k \leq h$) given by

\begin{equation}
\label{eq:update_hidden}
\sigma_k = {\rm sign} \left(\sum_{i=0}^N w_{ki}\xi_i\right) \equiv
{\rm sign} (\vec{w}_k\cdot{\vec\xi})
\end{equation}

\noindent define the pattern's $h$-dimensional IR, ${\vec\sigma}(h)=(1,\sigma_1,
 \dots , \sigma_h)$. The network's output $\zeta(h)$ is:

\begin{equation}
\label{eq:update_output}
\zeta(h) = {\rm sign} \left(\sum_{k=0}^h W_k \sigma_k\right) \equiv
{\rm sign} \left[ \vec{W}(h) \cdot \vec{\sigma}(h)  \right]
\end{equation}

\noindent Hereafter, $\vec{\sigma}^\mu(h) = (1,\sigma_1^\mu, \cdots, \sigma_h^\mu)$
is the $h$-dimensional IR associated by the network of $h$ hidden
units to pattern $\vec{\xi}^\mu$.  During the training process, $h$
increases through the addition of hidden neurons, and we denote $H$
the final number of hidden units.

\subsection {Example}

\begin{figure}
 \centering
 \includegraphics[height=6cm]{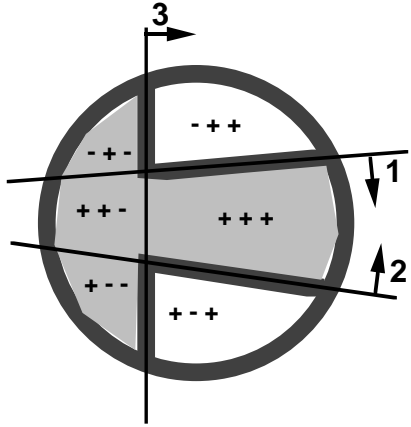}
\caption{Example: patterns inside the grey region belong to one
class, those in the white region to the other. The lines (labelled 1,
2 and 3) represent the hyperplanes found with the NetLines strategy.
The arrows point into the correspondent positive half-spaces. The IRs
of each domain are indicated (the first component, $\sigma_0=1$, is omitted for clarity).}
\label{fig:schema}
\end{figure}

Let us first describe the general strategy on a schematic example
(fig.~\ref{fig:schema}). Patterns in the grey region belong to
class $\tau=+1$, the others to $\tau=-1$. The algorithm proceeds as
follows: a first hidden unit is trained to separate the input
patterns at best, and finds one solution, say $\vec{w}_1$,
represented on fig.~\ref{fig:schema} by the line labelled $1$, with
the arrow pointing into the positive half-space. As there remain
training errors, a second hidden neuron is introduced. It is
trained to learn targets $\tau_2=+1$ for patterns well classified
by the first neuron, $\tau_2=-1$ for the others (the opposite
convention could be adopted, both being strictly equivalent), and
suppose that solution $\vec{w}_2$ is found. Then an output unit is
connected to the two hidden neurons and is trained with the
original targets. Clearly it will fail to separate correctly all
the patterns because the IR $(-1,1)$ is not faithful, as patterns
of both classes are mapped onto it.  The output neuron is dropped,
and a third hidden unit is appended and trained with targets
$\tau_3 =+1$ for patterns that were  correctly classified by the
output neuron and $\tau_3=-1$ for the others. Solution $\vec{w}_3$
is found, and it is easy to see that now the IRs are faithful, {\it
i.e.} patterns belonging to different classes are given different
IRs. The algorithm converged with 3 hidden units that define 3
domain boundaries determining 6 regions or domains in the input
space. It is straightforward to verify that the IRs correspondent
to each domain, indicated on fig.~\ref{fig:schema}, are linearly
separable. Thus, the output unit will find the correct solution to
the training problem. If the faithful IRs were not linearly
separable, the output unit would not find a solution without
training errors, and the algorithm would go on appending hidden
units that should learn targets $\tau=1$ for well learnt patterns,
and $\tau=-1$ for the others. A proof that a solution to this
strategy with a finite number of hidden units exists is left to the
Appendix.

\subsection{The Algorithm NetLines}
\label{subsec:NetLines}

Like most adaptive learning algorithms, NetLines combines a growth
heuristics with a particular learning algorithm for training the
individual units,  which are simple perceptrons. In this section we
present the growth heuristics first, followed by the description of
Minimerror, our perceptron learning algorithm.

Let us first introduce the following useful remark: if a neuron has
to learn a target $\tau$, and the learnt state turns out to be
$\sigma$, then the product $\sigma \tau=1$ if the target has been
correctly learnt, and  $\sigma \tau=-1$ otherwise.

Given a maximal accepted number of hidden units, $H_{max}$, and a
maximal  number of tolerated training errors, $E_{max}$, the
algorithm may be summarized as follows:\\

{\bf Algorithm NetLines}
\begin{itemize}

\item {\bf initialize} \\
      ${\bf h=0}$;\\
      {\bf set} the targets $\tau_{h+1}^{\mu}=\tau^{\mu}$ for $\mu=1,\cdots,P$;

\item {\bf repeat}
\label{train-h}
   \begin{enumerate} \item /*~train the hidden units~*/ \\
      ${\bf h = h+1}$; /*~connect hidden unit $h$ to the inputs~*/\\
      {\bf learn} the training set $\{\vec{\xi}^{\mu},\tau_h^{\mu}\}$, $\mu=1,\cdots,P$;\\
      after learning, $\sigma_h^{\mu}={\rm sign} (\vec{w}_h\cdot{\vec\xi}^{\mu})$, $\mu=1,\cdots,P$;\\
      {\bf if} $h=1$ /*~for the first hidden neuron~*/\\
         \begin{description}
         \item[ ] {\bf if} $\sigma_1^{\mu}=\tau_1^{\mu}$ $\forall \mu$ {\bf then stop.} /*~the training set is LS~*/;
         \item[ ]{\bf else} {\bf set} $\tau_{h+1}^{\mu}=\sigma_h^{\mu}
           \tau^{\mu}$ for $\mu=1,\cdots,P$; {\bf go to} 1;
         \end{description}
      {\bf end if}
   \item /*~learn the mapping between the IRs and the outputs~*/\\
      {\bf connect the output} neuron to the $h$ trained hidden units;\\
      {\bf learn} the training set $\{\vec{\sigma}^{\mu}(h),\tau^{\mu}\}$; $\mu=1,\cdots,P$;\\
      after learning, $\zeta^{\mu}(h)={\rm sign}\left(\vec{W}(h)\cdot \vec{\sigma}^{\mu}\right)$, $\mu=1,\cdots,P$;\\
      {\bf set} $\tau_{h+1}^{\mu}=\zeta^{\mu} \tau^{\mu}$ for $\mu=1,\cdots,P$;\\
      {\bf count} the number of training errors $e= \sum_{\mu} (1-\tau_{h+1}^{\mu})/2$;
    \end{enumerate}
\item {\bf until} $(h=H_{max}$ or $e\leq E_{max})$;

\end{itemize}

\noindent The generated network has $H=h$ hidden units. In the
Appendix we present a solution to the learning strategy with a
bounded number of hidden units. In practice the algorithm ends up
with much smaller networks than this upper bound, as will be shown
in Section \ref{sec:Benchmarks}.

\subsection{The perceptron learning algorithm}

The final number of hidden neurons, which are simple perceptrons,
depends on the performance of the learning algorithm used to train
them. The best solution should minimize the number of errors; if
the training set is LS it should endow the units with the lowest
generalization error. Our incremental algorithm uses Minimerror
\cite{Minimerror} to train the hidden and output units. Minimerror is
based on the minimization of a cost function $E$ that depends on the
perceptron weights $\vec w$ through the stabilities of the training
patterns. If the input vector is $\vec{\xi}^{\mu}$ and $\tau^{\mu}$
the corresponding target, then the {\it stability} $\gamma^{\mu}$ of
pattern $\mu$ is a continuous and derivable function of the weights,
given by:

\begin{equation}
\label{eq:stability}
\gamma^{\mu} = \tau^{\mu}
\frac{\vec w \cdot \vec{\xi}^{\mu}}{ \parallel \vec{w} \parallel }
\end{equation}

\noindent where $\parallel \vec{w} \parallel = \sqrt{\vec w \cdot
\vec w}$. The stability is independent of the norm of the
weights $\parallel \vec{w} \parallel$. It measures the distance of
the pattern to the separating hyperplane, which is normal to
$\vec{w}$; it is positive if the pattern is well classified,
negative otherwise. The cost function $E$ is:

\begin{equation}
\label{eq:cost}
E=\frac{1}{2} \sum_{\mu=1}^P \left[1- \tanh \frac{\gamma^{\mu}}{2T}\right]
\end{equation}

\noindent The contribution to $E$ of patterns with large negative
stabilities is $\simeq 1$, {\it i.e.} they are counted as errors,
whereas the contribution of patterns with large positive stabilities
is vanishingly small. Patterns at both sides of the hyperplane within
a window of width $\approx 4T$ contribute to the cost function even
if they have positive stability.

The properties of the global minimum of (\ref{eq:cost}) have been
studied theoretically with methods of statistical mechanics
\cite{GorGre}. It was shown that in the limit $T \rightarrow 0$, the
minimum of $E$ corresponds to the weights that minimize the number of
training errors. If the training set is LS, these weights are not
unique \cite{Gyorgyi90b}. In that case, there is an optimal
learning temperature such that the weights minimizing $E$ at that
temperature endow the perceptron with a generalization error
numerically indistinguishable from the optimal (bayesian) value.

The algorithm Minimerror \cite{Minimerror,nc:Raffin+Gordon:1995}
implements  a minimization of $E$ restricted to a sub-space of
normalized weights, through a gradient descent combined with a slow
decrease of the temperature $T$, which is equivalent to a
deterministic annealing. It has been shown that the convergence is
faster if patterns with negative stabilities are considered at a
temperature $T_-$ larger than those with positive stabilities,
$T_+$, with a constant ratio $\theta = T_-/T_+$. The weights and
the temperatures are iteratively updated through:

\begin{eqnarray}
\label{eq:Minimerror.1}
\delta \vec{w}(t)&=& \epsilon \left[\sum_{\mu/\gamma^\mu \leq 0}
\frac{\tau^\mu \vec{\xi}^\mu}{\cosh^2(\gamma^\mu/2T_-)} +
\sum_{\mu/\gamma^\mu > 0} \frac{\tau^\mu \vec{\xi}^\mu}
{\cosh^2(\gamma^\mu/2T_+)} \right]\\
\label{eq:Minimerror.2}
T_+^{-1}(t+1) &=& T_+^{-1}(t) + \delta T^{-1};\,\, T_-=\theta T_+;\\
\label{eq:Minimerror.3}
\vec{w}(t+1) &=& \sqrt{N+1}\frac{\vec{w}(t) + \delta
\vec{w}(t)}{\parallel \vec{w}(t) + \delta \vec{w}(t) \parallel }
\end{eqnarray}

\noindent
Notice from (\ref{eq:Minimerror.1}) that only the incorrectly learned patterns at distances shorter than $\approx 2T_-$ from the hyperplane, and those correctly learned lying closer than $\approx 2T_+$, contribute effectively to learning. The contribution of patterns outside this region are vanishingly small. By decreasing the temperature, the algorithm selects to learn patterns increasingly localized in the neighborhood of the hyperplane, allowing for a highly precise determination of the parameters defining the hyperplane, which are the neuron's weights. Normalization (\ref{eq:Minimerror.3}) restricts the search to the sub-space with $\parallel \vec{w} \parallel = \sqrt{N+1}$.

The only adjustable parameters of the algorithm are the temperature ratio $\theta=T_-/T_+$, the learning rate $\epsilon$ and the annealing rate $\delta T^{-1}$. In principle they should be adapted to each specific problem. However, thanks to our normalizing the weights to $\sqrt{N+1}$ and to data standardization (see next section), all the problems are brought to the same scale, simplifying the choice of the parameters.

\subsection{Data standardization}

Instead of determining the best parameters for each new problem, we
standardize the input patterns of the training set through a linear
transformation, applied to each component:

\begin{equation}
\label{eq:new_set}
\tilde{\xi}_i^{\mu}=\frac{\xi_i^{\mu} -
\langle\xi_i\rangle}{\Delta_i}; \; 1 \leq i \leq N
\end{equation}

\noindent The mean $\langle\xi_i\rangle$ and the variance
$\triangle_i^2$, defined as usual:

\begin{eqnarray}
\langle\xi_i\rangle & = & \frac{1}{P} \sum_{\mu=1}^P \xi_i^{\mu} \\
{\Delta_i}^2 & = & \frac{1}{P} \sum_{\mu=1}^P
{({\xi_i^{\mu} - \langle\xi_i\rangle})}^2 = \frac{1}{P} \sum_{\mu=1}^P
(\xi_i^{\mu})^2 - (\langle\xi_i\rangle)^2
\end{eqnarray}

\noindent need only a single pass of the $P$ training patterns to be
determined. After learning, the inverse transformation is applied to
the weights,

\begin{eqnarray}
\label{eq:new_bias}
\tilde{w}_0 &=& \sqrt{N+1} \frac{w_0 - \sum\limits_{i=1}^N w_i \langle
\xi_i\rangle/ \Delta_i}
{\sqrt{\left[ w_0 - \sum_{j=1}^N w_j \langle
\xi_j\rangle/ \Delta_j \right]^2 +
\sum_{j=1}^N (w_j /\Delta_j)^2}} \\
\label{eq:new_weights}
\tilde{w}_i &=& \sqrt{N+1} \frac{w_i / \Delta_i}
{\sqrt{\left[ w_0 - \sum_{j=1}^N w_j \langle \xi_j\rangle/
\Delta_j \right]^2 + \sum_{j=1}^N (w_j /\Delta_j)^2}}
\end{eqnarray}

\noindent so that the normalization (\ref{eq:new_set}) is completely
transparent to the user: with the transformed weights
(\ref{eq:new_bias}) and (\ref{eq:new_weights}) the neural classifier
is applied to the data in the original user's units, which do not
need to be renormalized.

As a consequence of the weights scaling (\ref{eq:Minimerror.3}) and
the inputs standardization (\ref{eq:new_set}), all the problems are
automatically rescaled. This allows us to use always the same values of Minimerror's parameters, namely, the ''standard'' values $\epsilon=0.02$, $\delta T^{-1}=10^{-3}$ and $\theta = 6$. They were used throughout this paper, the reported results being highly insensitive to slight variations of them. However, in some extremely difficult cases, like learning the parity in dimensions $N>10$ and finding the separation of the sonar signals (see section \ref{sec:Benchmarks}), larger values of $\theta$ were needed.

\subsection{Interpretation}
\label{sec:interpretation}

It has been shown \cite{GorPerBer} that the contribution of each
pattern to the cost function of Minimerror, $\left[1- \tanh
(\gamma^{\mu}/2T)\right]/2$, may be interpreted as the probability of
misclassification at the temperature $T$ at which the minimum of the
cost function has been determined. By analogy, the neuron's
prediction on a new input $\vec {\xi}$ may be given a confidence
measure by replacing the (unknown) pattern stability by its absolute
value $\parallel \gamma \parallel = \parallel \vec w \cdot
\vec{\xi}\parallel / \parallel \vec{w} \parallel$, which is its
distance to the hyperplane. This interpretation of the sigmoidal
function $\tanh (\parallel \gamma \parallel/2T)$ as the confidence on
the neuron's output is similar to the one proposed earlier
\cite{nc:Goodman+Smyth+Higgins:1992} within an approach based on
information theory.

The generalization of these ideas to multilayered networks is not
straightforward. An estimate of the confidence on the classification
by the output neuron should include the magnitude of the weighted
sums of the hidden neurons, as they measure the distances of the
input pattern to the domain boundaries. However, short distances to
the separating hyperplanes are not always correlated to low
confidence on the network's output. For an example, we refer again to
figure \ref{fig:schema}. Consider a pattern lying close to hyperplane
1. A small weighted sum on neuron 1 may cast doubt on the
classification if the pattern's IR is (-++), but not if it is (-+-),
as a change of the sign of the weighted sum in the latter case will
map the pattern to the IR (++-) which, being another IR of the same
class, will be given the same output by the network. It is worth noting that the same difficulty is met by the interpretation of the outputs of multilayered perceptrons, trained with Backpropagation, as posterior probabilities. 
We do not go on further into this problem, which is beyond the scope of this paper.

\section{Comparison with other strategies}
\label{sec:Comparison}

There are few learning algorithms for neural networks composed of
{\it binary} units. To our knowledge, all of them are incremental. In
this section we give a short overview of some of them, in order to
put forward the main differences with NetLines. We discuss the growth
heuristics first, and the individual units training algorithms
afterwards.

The Tiling algorithm \cite{tiling} introduces hidden layers, one after the other. 
The first neuron of each layer is trained to learn an IR that helps to decrease the number of training errors; supplementary hidden units are then appended to the layer until the IRs of all the patterns in the training set are faithful. 
This procedure may generate very large networks. The Upstart algorithm \cite{nc:Frean:1990} introduces successive couples of daughter hidden units between the input layer and the previously included hidden units, which become their parents. 
The daughters are trained to correct the parents classification errors, one daughter for each class. The obtained network has a tree-like architecture. 
There are two different algorithms implementing the Tilinglike Learning in the Parity Machine \cite{Tiling-Parity}, Offset
\cite{Offset} and MonoPlane \cite{Monoplane}. In both algorithms,
each appended unit is trained to correct the errors of the
previously included unit in the same hidden layer, a procedure that has been shown to generate a {\it parity machine}: 
the class of the input patterns is the parity of the learnt IRs. Unlike Offset, which implements the parity through a second hidden layer, that needs to be pruned,
MonoPlane goes on adding hidden units (if necessary) in the same
hidden layer until the number of training errors at the output vanishes.
Convergence proofs for {\it binary input patterns} have been produced
for all these algorithms. In the case of {\it real-valued input
patterns}, a solution to the parity machine with a bounded number of
hidden units also exists \cite{Convergence}.

The rationale behind the construction of the parity machine is that
it is not worth training the output unit before all the training
errors of the hidden units have been corrected. However, Marchand
{\it et al.} \cite{Marchand90} pointed out that it is not necessary
to correct {\it all} the errors of the successively trained hidden
units: it is sufficient that the IRs be faithful and LS. If the
output  unit is trained immediately after each appended hidden unit,
the network may discover that the IRs are already faithful and stop
adding units. This may be seen on the example of figure
\ref{fig:schema}. None of the parity machine implementations would
find the solution represented on the figure, as each of the 3
perceptrons unlearns systematically part of the patterns learnt by
the preceding one.

To our knowledge, Sequential Learning \cite{Marchand90} is the only
incremental learning algorithm that might find a solution equivalent
(although not the same) to the one of figure \ref{fig:schema}. In
this algorithm, the first unit is trained to separate the training
set keeping one ''pure'' half-space, {\it i.e.} a half space only
containing patterns of one class. Wrongly classified patterns, if
any, must all lie in the other half-space. Each appended neuron is
trained to separate wrongly classified patterns with this constraint,
{\it i.e.} keeping always one ''pure'', error-free, half-space. Thus,
neurons must be appended in a precise order, making the algorithm difficult to implement in practice. 
For example, Sequential Learning applied to the problem of figure \ref{fig:schema} needs to
impose that the first unit finds the weights $\vec{w}_3$, as this is
the only solution satisfying the purity restriction.

Other proposed incremental learning algorithms strive to solve the
problem with different architectures, and/or with real valued units.
For example, in the algorithm Cascade Correlation~\cite{Fahlman90},
each appended unit is selected among a pool of several real-valued
neurons, trained to learn the correlation between the targets and the
training errors. The unit is then connected to the input units and to
all the other hidden neurons already included in the network.

Another approach to learning classification tasks is through the
construction of decision trees \cite{Breiman+Friedman+Olshen:1984},
which partition hierarchically the input space through successive
dichotomies. The neural networks implementations generate tree-like
architectures. Each neuron
of the tree introduces a dichotomy of the input space which is
treated separately by the children nodes, which eventually produce
new splits. Besides the weights, the resulting networks need to
store the decision path. The proposed heuristics
\cite{neuraltree,nips-6:Farrell+Mammone:1994,Stepwise} differ in the
algorithm used to train each node, and/or in the stopping criterium.
In particular, Neural-Trees \cite{neuraltree} may be regarded as a
generalization of CART \cite{Breiman+Friedman+Olshen:1984} in which
the hyperplanes are not constrained to be perpendicular to the
coordinate axis. The heuristics of the Modified Neural Tree Network
\cite{nips-6:Farrell+Mammone:1994}, similar to Neural-Trees,
includes a criterium of early stopping based on a confidence
measure of the partition. As NetLines considers the whole input
space to train each hidden unit, it generates domain boundaries
which may greatly differ from the splits produced by trees.
We are not aware of any systematic study nor theoretical comparison
of both approaches.

Other algorithms, like RCE \cite{RCE}, GAL \cite{Gal}, Glocal
\cite{Glocal} and Growing cells \cite{nips-6:Fritzke:1994} propose to
cover or mask the input space with hyperspheres of adaptive size
containing patterns of the same class. These approaches generally end
up with a very large number of units. Covering Regions by the LP
Method \cite{nc:Mukhopadhyay+Roy+Kim:1993} is a trial and error
procedure devised to select the most efficient masks among
hyperplanes, hyperspheres or hyperellipsoids. The mask's parameters
are determined through linear programming.

Many incremental strategies use the Pocket algorithm \cite{Pocket} to
train the appended units. Its main drawback is that it has no natural
stopping condition, which is left to the user's patience. The
proposed alternative algorithms
\cite{nc:Frean:1992,nc:Bottou+Vapnik:1992} are not guaranteed to find
the best solution to the problem of learning. The algorithm used by
the Modified Neural Tree Network (MNTN)
\cite{nips-6:Farrell+Mammone:1994} and the ITRULE
\cite{nc:Goodman+Smyth+Higgins:1992} minimize cost functions similar
to (\ref{eq:cost}), but using different misclassification measures at
the place of our stability (\ref{eq:stability}). The {\it essential}
difference with Minimerror is that none of these algorithms are able to
control which patterns contribute to learning, like Minimerror does
with the temperature.

\section{Generalization to Multi-class Problems}
\label{sec:Multiclass}

The usual way to cope with problems having more than two classes, is
to generate as many networks as classes. Each network is trained to
separate patterns of one class from all the others, and a
winner-takes-all (WTA) strategy based on the value of output's
weighted sum in equation (\ref{eq:update_output}) is used to decide
the class if more than one network recognizes the input pattern. As
we use normalized weights, in our case the output's weighted sum is
merely the distance of the IR to the separating hyperplane. All the
patterns mapped to the same IR are given the same output's weighted
sum, independently of the relative position of the pattern in input
space. As already discussed in section \ref{sec:interpretation},
a strong weighted sum on the {\it output neuron} is not
inconsistent with small weighted sums on the hidden neurons.
Therefore, a naive WTA decision may not give good results, as is
shown in the example of section \ref{sec:Breiman}.

We now describe an implementation for the multi-class problem that
results in a tree-like architecture of {\it networks}. It is more
involved that the naive WTA, and may be applied to any binary
classifier. Suppose that we have a problem with $C$ classes. We
must choose in which order the classes will be learnt, say $(c_1,
c_2,
\cdots, c_C)$. This order constitutes a particular {\it learning
sequence}. Given a particular learning sequence, a first network is
trained to separate class $c_1$, which is given output target
$\tau_1=+1$, from the others (which are given targets $\tau_1=-1$).
The opposite convention is equivalent, and could equally be used.
After training, all the patterns of class $c_1$ are eliminated from
the training set and we generate a second network trained to separate
patterns of class $c_2$ from the remaining classes. The procedure,
reiterated with training sets of decreasing size, generates $C-1$
hierarchically organized tree of networks (TON): the outputs are
ordered sequences $\vec\zeta= (\zeta_1, \zeta_2, \cdots, \zeta_{C-1})$. The
predicted class of a pattern is $c_i$, where $i$ is the first network
in the sequence having an output $+1$ ($\zeta_i=+1$ and $\zeta_j=-1$
for $j<i$), the outputs of the networks with $j>i$ being irrelevant.

The performance of the TON may depend on the chosen learning
sequence. Therefore, it is convenient that an odd number of TONs,
trained with different learning sequences, compete through a vote. We
verified empirically, as is shown in section \ref{sec:Multiclass
problems}, that this vote improves the results obtained with each of
the individual TONs participating to the vote. Notice that our
procedure is different from bagging \cite{Bagging_predictors}, as all
the networks of the TON are trained with the {\it same} training set,
without the need of any resampling procedure.

\section{Applications}
\label{sec:Benchmarks}

Although convergence proofs of learning algorithms are satisfactory
on theoretical grounds, they are not a guarantee of good
generalization. In fact, they only demonstrate that correct {\it
learning} is possible, but do not address the problem of
generalization. This last issue still remains quite empirical
\cite{nips-4:Vapnik:1992,nc:Geman+Bienenstock+Doursat:1992,Bias_Variance_Curse},
and the generalization performance of learning algorithms is usually
tested on well known benchmarks \cite{PrecheltL1994d}.

We first tested the algorithm on learning the parity function of $N$
bits for $2\leq N \leq 11$. It is well known that the smallest
network with the architecture considered here needs $H=N$ hidden
neurons. The optimal architecture was found in all the cases.
Although this is quite an unusual performance, the parity is not a
representative problem: learning  is exhaustive and generalization
cannot be tested. Another test, the classification of sonar signals
\cite{Gorman88a}, revealed the quality of Minimerror, as it solved
the problem without hidden units. In fact, we found that not only the
training set of this benchmark is linearly separable, a result
already reported \cite{Hoehfeld,nn:sonar_lineal:1993}, but that the
complete data base, {\it i.e.} the training and the test sets
together, are also linearly separable.

We present next our results, generalization error $\epsilon_g$ and
number of weights, on several benchmarks corresponding to different
kinds of problems: binary classification of binary input patterns,
binary classification of real-valued input patterns, and
multi-class problems. These benchmarks were chosen because they
served already as a test to many other algorithms, providing
unbiased results to compare with. The generalization error
$\epsilon_g$ of NetLines was estimated as usual, through the
fraction of misclassified patterns on a test set of data.

The results are reported as a function of the training sets sizes $P$
whenever these sizes are not specified by the benchmark. Besides the
generalization error $\epsilon_g$, averaged over a (specified) number
of classifiers trained with randomly selected training sets, we
present also the number of weights of the corresponding networks. The
latter is a measure of the classifier's complexity, as it corresponds
to the number of its parameters.

Training times are usually cited among the characteristics of the
training algorithms. Only the numbers of epochs used by
Backpropagation on two of the studied benchmarks have been published;
we restrict the comparison to these cases. As NetLines only updates
$N$ weights per epoch, whereas Backpropagation updates all the
network's weights, we compare the {\it total} number of weights
updates. They are of the same order of magnitude for both algorithms.
However, these comparisons should be taken with caution. NetLines is
a deterministic algorithm: it learns the architecture and the weights
through one single run, whereas with Backpropagation several
architectures must be previously investigated, and this time is not
included in the training time.

The following notation is used: $D$ is the total number of available
patterns, $P$ the number of training patterns, $G$ the number of test
patterns.

\subsection{Binary inputs}

The case of binary input patterns has the property, not shared by
real-valued inputs, that every pattern may be separated from
the others by a single hyperplane. This solution, usually called {\it grand-mother}, needs as many hidden units as patterns in the training set. 
In fact, the convergence proofs for incremental algorithms in the case of binary input patterns are based on this property.

\subsubsection{Monk's problem}

This benchmark, thoroughly studied with many different learning
algorithms \cite{Monks}, contains three distinct problems. Each one
has an underlying logical proposition that depends on six discrete
variables, coded with $N=17$ binary numbers. The total number of
possible input patterns is $D=432$, and the targets correspond to the
truth table of the corresponding proposition. Both NetLines and
MonoPlane found the underlying logical proposition of the first two
problems, {\it i.e.} they generalized correctly giving $\epsilon_g =
0$. In fact, these are easy problems: all the neural network-based
algorithms, and some non-neural learning algorithms were reported to
correctly generalize them. In the third Monk's problem, 6 patterns
among the $P_3=122$ examples are given wrong targets. The
generalization error is calculated over the complete set of $D=432$
patterns, {\it i.e.} including the training patterns, but in the test
set all the patterns are given the correct targets. Thus, any
training method that learns correctly the training set will make at
least $1.4\%$ of generalization errors. Four algorithms specially
adapted to noisy problems were reported to reach $\epsilon_g=0$. However, none of them generalizes correctly the two other (noiseless) Monk's problems. 
Besides them, the best performance, $\epsilon_g=0.0277$ which corresponds to 12 misclassified patterns, is only reached by neural networks methods: 
Backpropagation, Backpropagation with Weight Decay, Cascade-Correlation and NetLines. 
The number of hidden units generated with NetLines ($58$ weights) is intermediate between
Backpropagation with weight decay ($39$), and Cascade-Correlation
($75$) or Backpropagation ($77$). MonoPlane reached a slightly worse
performance ($\epsilon_g=0.0416$, {\it i.e.} 18 misclassified
patterns) with the same number of weights as NetLines, showing that
the parity machine encoding may not be optimal.

\subsubsection{Two or more clumps}

\begin{figure}
 \centering
 \includegraphics[height=6.5cm]{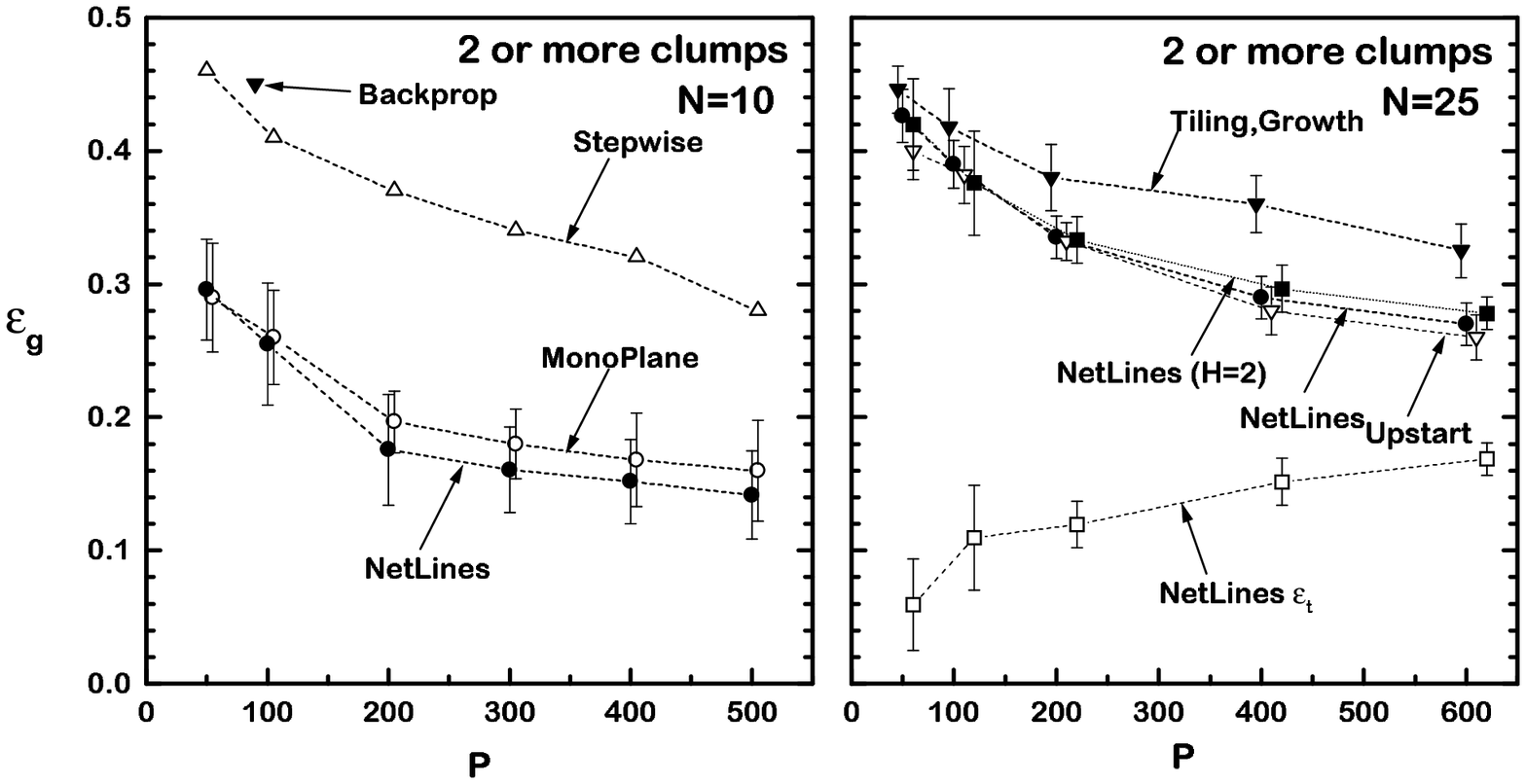}
\caption{Two or more clumps for two ring sizes, $N=10$ and $N=25$.
Generalization error $\epsilon_g$ vs. size of the training set $P$,
for different algorithms. $N=10$: Backpropagation
\protect{\cite{Solla89}}, Stepwise
\protect{\cite{Stepwise}}. $N=25$: Tiling
\protect{\cite{tiling}}, Upstart
\protect{\cite{nc:Frean:1990}}. Results with the Growth Algorithm
\protect{\cite{jpn_ijns}} are indistinguishable from those of
Tiling at the scale of the figure. Points without error bars
correspond to best results. Results of MonoPlane and
NetLines are averages over 25 tests.}
\label{fig:clumps-g}
\end{figure}

\begin{figure}
 \centering
 \includegraphics[height=6.5cm]{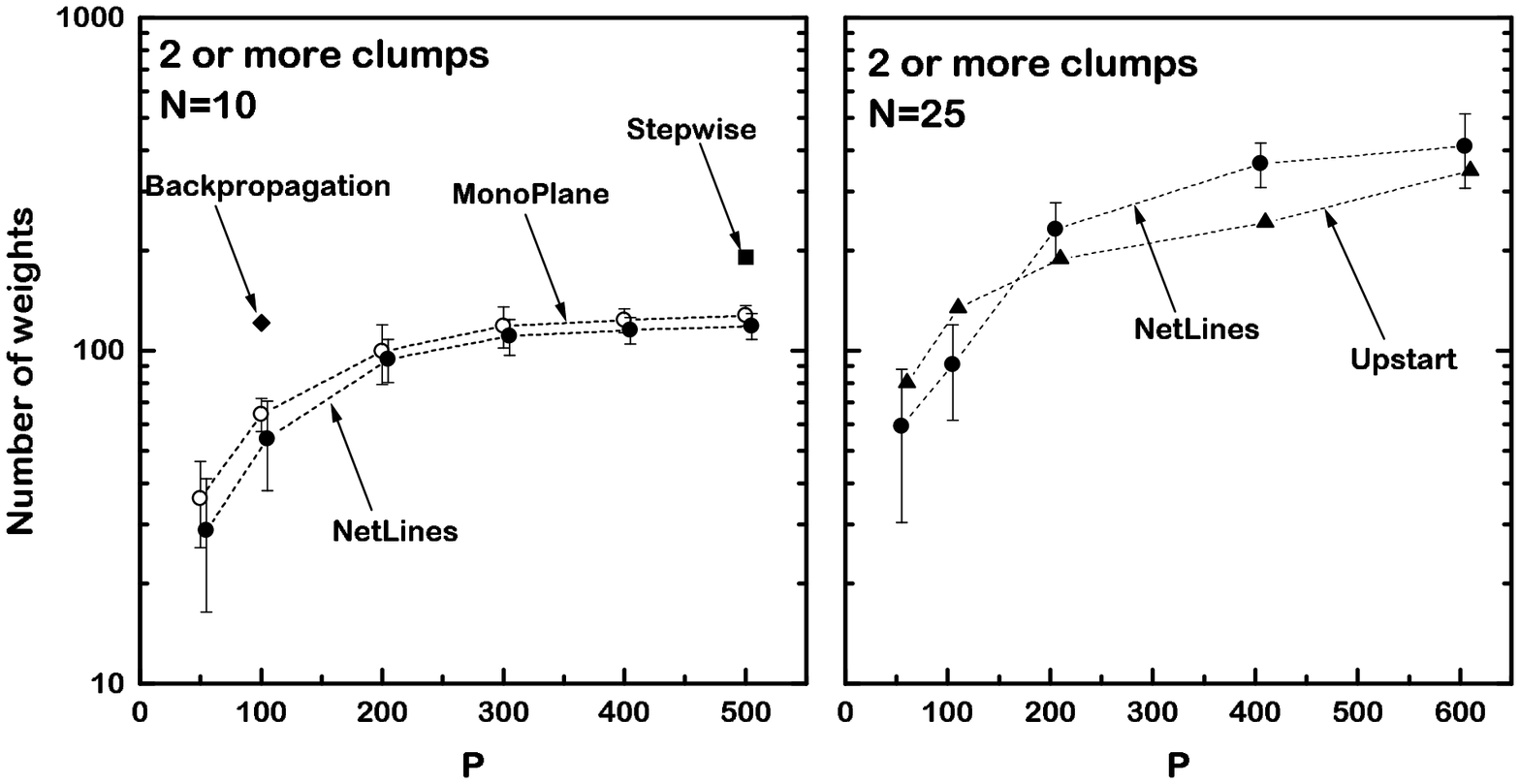}
  \caption{Two or more clumps. Number of weights (logarithmic scale)
  vs. size of the training set $P$, for $N=10$ and $N=25$. Results of
  MonoPlane and NetLines are averages over 25 tests. The references
  are the same as in figure \protect{\ref{fig:clumps-g}.}
  }
  \label{fig:clumps-w}
\end{figure}

In this problem \cite{Denker87} the network has to discriminate if
the number of clumps in a ring of $N$ bits is strictly smaller than 2
or not. One clump is a sequence of identical bits bounded by bits of
the other kind. The patterns are generated through a MonteCarlo
method in which the mean number of clumps is controlled by a
parameter $k$ \cite{tiling}. We generated training sets of $P$
patterns with $k=3$, corresponding to a mean number of clumps of
$\approx 1.5$, for rings of $N=10$ and $N=25$ bits. The
generalization error corresponding to several learning algorithms,
estimated with independently generated testing sets of the same sizes
as the training sets, {\it i.e.} $G=P$, are displayed on figures
\ref{fig:clumps-g} as a function of $P$. Points with error bars
correspond to averages over 25 independent training sets. Points
without error bars correspond to {\it best} results. NetLines,
MonoPlane and Upstart for $N=25$ have nearly the same performances
when trained to reach error-free learning.

We tested the effect of {\it early stopping} by imposing to
NetLines a  maximal number of two hidden units ($H = 2$). The
residual training error $\epsilon_t$ is plotted on the same figure,
as a function of $P$. It may be seen that early-stopping does not
help to decrease $\epsilon_g$. Overfitting, that arises when
NetLines is applied until error-free training is reached, does not
degrade the network's generalization performance. This behavior is
very different from the one of networks trained with
Backpropagation. The latter reduces classification learning to a
regression problem, in which the generalization error can be
decomposed in two competing terms: bias and variance. With
Backpropagation, early stopping helps to decrease overfitting
because some hidden neurons do not reach large enough weights to
work in the non-linear part of the sigmoidal transfer functions. It
is well known that all the neurons working in the linear part may
be replaced by a single linear unit. Thus, with early-stopping, the
network is equivalent to a smaller one, {\it i.e.} having less
parameters, with all the units working in the non-linear regime.
Our results are consistent with recent theories
\cite{Bias_Variance_Curse} showing that, contrary to regression,
the bias and variance components of the generalization error in
classification combine in a highly non-linear way.

The number of weights used by the different algorithms is plotted on
a logarithmic scale as a function of $P$ on figures
\ref{fig:clumps-w}. It turns out that the strategy of NetLines is
slightly better than that of MonoPlane, with respect to both
generalization performance and network size.

\subsection{Real valued inputs}

We tested NetLines on two problems which have real valued inputs (we
include graded-valued inputs here).

\subsubsection{Wisconsin Breast Cancer Data Base}

\begin{figure}
 \centering
 \includegraphics[height=6.5cm]{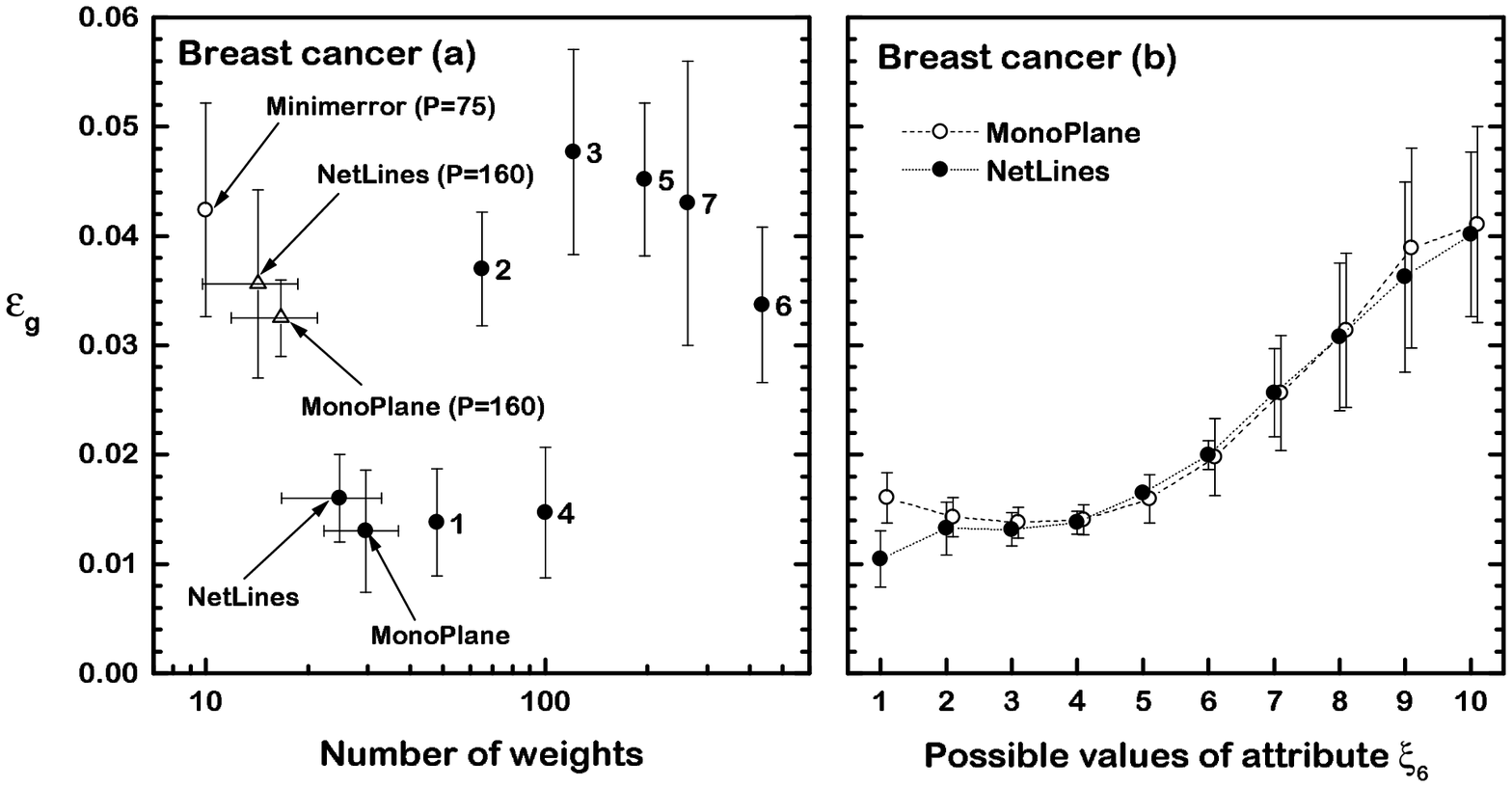}
\caption{Breast cancer classification. (a) Generalization error
$\epsilon_g$ vs. number of weights (logarithmic scale), for $P=525$.
1, 2, 3: Rprop with no shortcuts \protect{\cite{PrecheltL1994d}}; 4,
5, 6: Rprop with shortcuts \protect{\cite{PrecheltL1994d}}; 7:
Cascade Correlation \protect{\cite{Glocal}}. For comparison, results
with smaller training sets, $P=75$ (single perceptron) and $P=160$,
patterns are displayed. Results of MonoPlane and NetLines are
averages over 50 tests. (b) Classification errors vs. possible values
of the missing attribute ''{\it bare nuclei}'' for the 16 incomplete
patterns, averaged over 50 independently trained networks.}
\label{fig:cancer-gw}
\end{figure}

The input patterns of this benchmark~\cite{Breast_cancer} have
$N=9$ attributes characterizing samples of breast cytology,
classified as benign or malignant. We excluded from the original
data base 16 patterns that have the attribute $\xi_6$ ("{\it bare
nuclei}") missing. Among the remaining $D=683$ patterns, the two
classes are unevenly represented, 65.5\% of the examples being
benign. We studied the generalization performance of networks
trained with sets of several sizes $P$. The $P$ patterns for each
learning test were selected at random. On figure
~\ref{fig:cancer-gw}a, the generalization error at classifying the
remaining $G \equiv D-P$ patterns is displayed as a function of the
corresponding number of weights in a logarithmic scale. For
comparison, we included in the same figure results of a single
perceptron trained with $P=75$ patterns using Minimerror. The
results, averaged values over 50 independent tests for each $P$,
show that both NetLines and MonoPlane have lower $\epsilon_g$ and
less number of parameters than other algorithms on this benchmark.

The total number of weights updates needed by NetLines, including
the weights of the dropped output units, is $7\cdot10^4$;
Backpropagation needed $\approx 10^4$ \cite{PrecheltL1994d}.

The trained network may be used to classify the patterns with {\it
missing attributes}. The number of misclassified patterns among the
16 cases for which attribute $\xi_6$ is missing, is plotted as a
function of the possible values of $\xi_6$ on figure
~\ref{fig:cancer-gw}b. For large values of $\xi_6$ there are
discrepancies between the medical and the network's diagnosis on half
the cases. This is an example of the kind of information that may be
obtained in practical applications.

\subsubsection{Diabetes diagnosis}

\begin{figure}
 \centering
 \includegraphics[height=6cm]{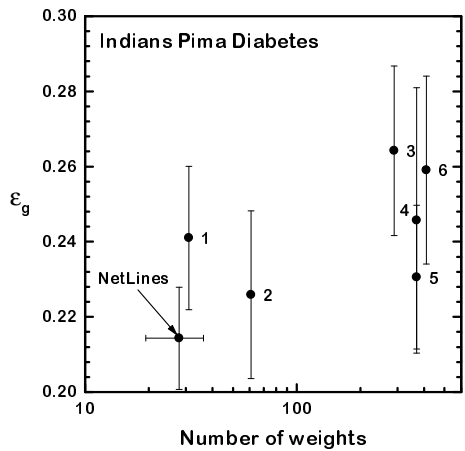}
\caption{Diabetes diagnosis: Generalization error $\epsilon_g$
vs. number of weights.
Results of NetLines are averages over $50$ tests.
1,2,3: Rprop no shortcuts, 4,5,6: Rprop with shortcuts
\protect{\cite{PrecheltL1994d}}
}
\label{fig:pima-gw}
\end{figure}

This benchmark~\cite{PrecheltL1994d} contains $D=768$ patterns
described by $N=8$ real-valued attributes, corresponding to $\approx
35\%$ of Pima women suffering from diabetes, $65\%$ being healthy.
Training sets of $P=576$ patterns were selected at random, and
generalization was tested on the remaining $G=192$ patterns. The
comparison with published results obtained with other algorithms tested
under the same conditions, presented on figure ~\ref{fig:pima-gw},
shows that NetLines reaches the best performances published so far
on this benchmark, needing much less parameters. Training times of
NetLines are of $\approx 10^5$ updates. The numbers of updates needed
by Rprop \cite{PrecheltL1994d} range between $4\cdot10^3$ and
$5\cdot10^5$, depending on the network's architecture.

\subsection{Multi-class problems}
\label{sec:Multiclass problems}

We applied our learning algorithm to two different problems, both of
three classes. We compare the results obtained with a
winner-takes-all (WTA) classification based on the results of three
networks, each one independently trained to separate one class from
the two others, to the results of the TON architectures described in
section \ref{sec:Multiclass}. As the number of classes is low, we
determined the three TONs, corresponding to the three possible
learning sequences. The vote of the three TONs improves the
performances, as expected.

\subsubsection{Breiman's Waveform Recognition Problem}
\label{sec:Breiman}

This problem was introduced as a test for the algorithm CART
\cite{Breiman+Friedman+Olshen:1984}. The input patterns are defined
by $N=21$ real-valued amplitudes $x(t)$ observed at regularly spaced
intervals $t=1,2, \cdots,N$. Each pattern is a noisy convex linear
combination of two among three elementary waves (triangular waves
centered on three different values of $t$). There are three possible
combinations, and the pattern's class identifies from which
combination it is issued.

\begin{figure}
 \centering
 \includegraphics[height=6cm]{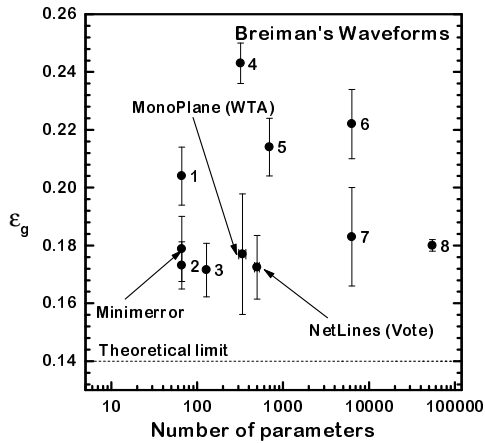}
\caption{Breiman waveforms: Generalization error $\epsilon_g$
averaged over 11 tests vs. number of parameters. Error bars on the
number of weights generated by NetLines and MonoPlane are not
visible at the scale of the figure. 1: Linear disc., 2: Perceptron,
3: Backpropagation, 4: Genetic algorithm, 5: Quadratic disc., 6:
Parzen's kernel. 7: K-NN, 8: Constraint \protect{\cite{Symenu}} }
\label{fig:Breiman-gw}
\end{figure}

We trained the networks with the same $11$ training sets of $P=300$
examples, and generalization was tested on the same independent test
set of $G=5000$, as in \cite{Symenu}. Our results are displayed on
figure \ref{fig:Breiman-gw}, where only results of algorithms
reaching $\epsilon_g < 0.25$ in \cite{Symenu} are included. Although
it is known that, due to the noise, the classification error has a
lower bound of $\approx 14\%$ \cite{Breiman+Friedman+Olshen:1984},
the results of NetLines and MonoPlane presented here correspond to
error-free training. The networks generated by NetLines have between
3 and 6 hidden neurons, depending on the training sets. The results
obtained with a single perceptron trained with Minimerror and with
the Perceptron learning algorithm, which may be considered as the
extreme case of early stopping, are hardly improved by the more
complex networks. Here again the overfitting produced by error-free
learning with NetLines does not deteriorate the generalization
performance. The TONs vote reduces the variance, but does not
decrease the average $\epsilon_g$.

\subsubsection{Fisher's Iris plants database}

\begin{figure}
 \centering
 \includegraphics[height=6cm]{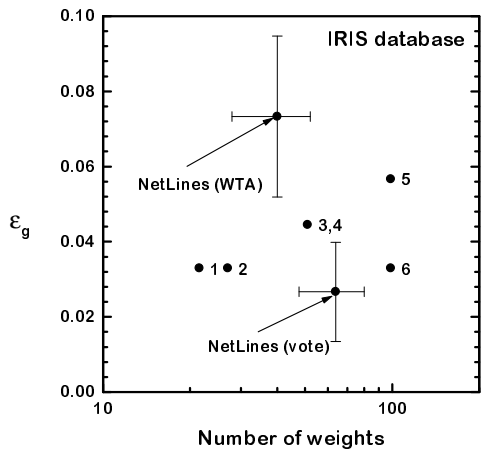}
\caption{Iris database: Generalization error $\epsilon_g$ vs.
number of parameters. 1: Offset, 2: Backpropagation
\protect{\cite{Offset}}; 4,5: Backpropagation \protect{\cite{Got}}; 3,6: GOT \protect{\cite{Got}}.
}
\label{fig:iris-gw}
\end{figure}

In this classical three-class problem, one has to determine
the class of Iris plants, based on the values of $N=4$ real-valued
attributes. The database of $D=150$ patterns, contains $50$ examples
of each class. Networks were trained with $P=149$ patterns, and the
generalization error is the mean value of all the $150$ leave-one-out
possible tests. Results of $\epsilon_g$ are displayed as a function
of the number of weights on figure \ref{fig:iris-gw}. Error bars are
available only for our own results. In this hard problem, the vote of
the three possible TONs trained with the three possible
class sequences (see section \ref{sec:Multiclass}) improves the
generalization performance.

\section{Conclusion}
\label{sec:conclusion}

We presented an incremental learning algorithm for classification,
that we call NetLines for {\bf N}eural {\bf E}ncoder {\bf T}hrough
{\bf Line}ar {\bf S}eparation. It generates small feedforward
neural networks with a single hidden layer of binary units
connected to a binary output neuron. NetLines allows for an
automatic adaptation of the neural network to the complexity of the
particular task. This is  achieved by coupling an error correcting
strategy for the successive addition of hidden neurons with
Minimerror, a very efficient perceptron training algorithm.
Learning is fast, not only because it reduces the problem to that
of training single perceptrons, but mainly because there is no
longer need of the usual preliminary tests necessary to determine
the correct architecture for the particular application. Theorems
valid for binary as well as for real-valued inputs guarantee the
existence of a solution with a bounded number of hidden neurons
obeying the growth strategy.

The networks are composed of binary hidden units whose states
constitute a faithful encoding of the input patterns. They
implement a mapping from the input space to a discrete
$H$-dimensional hidden space, $H$ being the number of hidden
neurons. Thus, each pattern is  labelled with a binary word of $H$
bits. This encoding may be seen as a compression of the pattern's
information. The hidden neurons define linear boundaries, or
portions of boundaries, between classes in input space. The
network's output may be given a probabilistic interpretation based
on the distance of the patterns to these boundaries.

Tests on several benchmarks showed that the networks generated  by
our incremental strategy are {\it small}, in spite of the fact that
the hidden neurons are appended until error-free learning is
reached. Even in those cases where the networks obtained with
NetLines are larger than those used by other algorithms, its
generalization error remains among the smallest values reported. In
noisy or difficult problems it may be useful to stop the network's
growth before the condition of zero training errors is reached.
This  decreases overfitting, as smaller networks (with less
parameters) are thus generated. However, the prediction quality
(measured by the generalization error) of the classifiers generated
with NetLines are not improved by early-stopping.

The results presented in this paper were obtained without
cross-validation,  nor any data manipulation like boosting, bagging
or arcing
\cite{Bagging_predictors,nips-5:Drucker+Schapire+Simard:1993}.
Those costly procedures combine results of very large numbers of
classifiers, with the aim of improving the generalization
performance through the reduction of the variance. As NetLines is a
stable classifier, presenting small variance, we do not expect that
such techniques would significantly improve our results.


\section*{Appendix}

In this Appendix we exhibit a particular solution to the learning
strategy of NetLines. This solution is built in such a way that the
cardinal of a convex subset of well learnt patterns, $L_h$, grows
monotonically upon the addition of hidden units. As this cardinal
cannot be larger that the total number of training patterns, the
algorithm must stop with a finite number of hidden units.

Suppose that $h$ hidden units have already been included
and that the output neuron still makes classification errors on
patterns of the training set, called training errors. Among these
wrongly learned patterns, be $\nu$ the one closest
to the hyperplane normal to $\vec{w}_h$, called hyperplane-$h$
hereafter. We define $L_h$ as the subset of (correctly learnt)
patterns laying closer to the hyperplane-$h$ than
$\vec{\xi}^{\nu}$. Patterns in $L_h$ have $0 < \gamma_h <
|\gamma_h^{\nu}|$. The subset $L_h$ and at least pattern $\nu$ are
well learnt if the next hidden unit, $h+1$, has weights:

\begin{equation}
\label{eq:hidden_weights}
\vec w_{h+1}=\tau_h^{\nu} \vec w_{h} - (1-\epsilon_h)
\tau_h^{\nu} (\vec w_{h} \cdot \vec{\xi}^{\nu}) \hat{e}_0
\end{equation}

\noindent where $\hat{e}_0 \equiv (1,0, \cdots, 0)$. The conditions
that both $L_h$ and pattern $\nu$ have positive stabilities ({\it
i.e.} be correctly learned) impose that

\begin{equation}
0 < \epsilon_h < \min_{\mu\in L_h}
\frac{|\gamma_h^{\nu}|-\gamma_h^\mu}{|\gamma_h^{\nu}|}
\end{equation}

The following weights between the hidden units and the output will
give the correct output to pattern $\nu$ {\it and} to the patterns of
$L_h$:

\begin{eqnarray}
W_0(h+1) & = & W_0(h) + \tau^\nu \\
W_i(h+1) & = & W_i(h) \; for \; 1 \leq i \leq h\\
W_{h+1}(h+1) & = & -\tau^\nu
\end{eqnarray}

\noindent Thus, $card(L_{h+1}) \geq card(L_h) +1$. As the number of
patterns in $L_h$ increases monotonically with $h$, convergence is
guaranteed with less than $P$ hidden units.

\end{document}